\documentclass[11pt,a4paper]{article}
\usepackage[hyperref]{acl2021}
\usepackage{times}
\usepackage{latexsym}

\usepackage{microtype}

\usepackage{times}
\usepackage{latexsym}
\usepackage{amsmath}
\usepackage{graphicx}
\usepackage{subfigure}
\usepackage{color}

\aclfinalcopy 
\def\aclpaperid{1278} 

\title{Improving Gradient-based Adversarial Training for Text Classification by Contrastive Learning and Auto-Encoder}

\author{Yao Qiu, \ Jinchao Zhang, \ Jie Zhou \\
  Pattern Recognition Center, WeChat AI, Tencent Inc, China \\
  \texttt{\{yasinqiu, dayerzhang, withtomzhou\}@tencent.com}
}
  
\date{}

\begin{document}

\maketitle
\begin{abstract}
Recent work has proposed several efficient approaches for generating gradient-based adversarial perturbations on embeddings and proved that the model's performance and robustness can be improved when they are trained with these contaminated embeddings. While they paid little attention to how to help the model to learn these adversarial samples more efficiently.
In this work, we focus on enhancing the model's ability to defend gradient-based adversarial attack during the model's training process and propose two novel adversarial training approaches: (1) CARL narrows the original sample and its adversarial sample in the representation space while enlarging their distance from different labeled samples. (2) RAR forces the model to reconstruct the original sample from its adversarial representation.
Experiments show that the proposed two approaches outperform strong baselines on various text classification datasets. Analysis experiments find that when using our approaches, the semantic representation of the input sentence won't be significantly affected by adversarial perturbations, and the model's performance drops less under adversarial attack. That is to say, our approaches can effectively improve the robustness of the model. Besides, RAR can also be used to generate text-form adversarial samples.
\end{abstract}

\section{Introduction}
Text classification is a fundamental research topic in natural language processing \cite{pang2002thumbs, lai2015recurrent, neekhara2018adversarial, sun2019fine}. Neural networks have obtained state-of-the-art performance on many text classification datasets \cite{kim2014convolutional, wang2018glue, devlin2018bert}. Despite these models' success, recent work has shown that they can be easily fooled by intentionally designed adversarial examples. These adversarial examples generated by adding little perturbations on original examples cannot affect human's judgment but can fail models \cite{ren2019generating, xu2019lexicalat}.

Adversarial training approaches are proposed to tackle this problem, which aims to enhance the model's strength of generalization and robustness by generating adversarial samples and letting the model learn them \cite{ren-etal-2019-generating, xu2019lexicalat}.
The approaches for generating adversarial samples can be roughly classified into two categories: text-based and gradient-based. The former can be further classified into three levels: character-level, word-level, and sentence-level. Compared to gradient-based adversarial approaches, the text-based are explainable, but they may suffer from low attack diversity and rely more on human knowledge which limits the kinds of adversarial patterns. In contrast, during the gradient-based adversarial training process, small perturbations calculated from the gradient are added to mini-batches embeddings of original training samples, then the model's parameters will be optimized to correctly classify the original embeddings together with adversarial embeddings \cite{miyato2016adversarial}. This kind of approach consists of two major steps: adversarial perturbation's construction and adversarial sample's learning.
Recent approaches mainly focus on the first step, as for the second step, only the classification loss is used by the model to learn the adversarial samples.

In this work, we investigate gradient-based adversarial training and focus on the second step. To further improve model's robustness against adversarial perturbations, we propose two approaches for text classification models: CARL (\textbf{C}ontrastive \textbf{A}dversarial \textbf{R}epresentation \textbf{L}earning) and RAR (\textbf{R}econstruction from \textbf{A}dversarial \textbf{R}epresentations).
We first generate adversarial samples by adding perturbations on input sentence's word embeddings, then CARL and RAR are used to learn these adversarial samples.  
CARL leverages the family of contrastive objectives \cite{gutmann2010noise, hjelm2018learning, tian2019contrastive} and aims to prevent the semantic representation of input sentence from being affected by adversarial attacks by narrowing the distance between the adversarial sample and its corresponding original sample in the representation space, while pushing them apart from samples which belong to different classes. If the representations of adversarial sample and original sample are 
identical, the model won't be fragile to the adversarial attack.
While CARL's goal is to learn a robust sentence-level representation, RAR acts like an auto-encoder and is designed to improve the robustness of the representation for each word by forcing the model to reconstruct original words from their adversarial embeddings. It will be much easier for the model to understand the adversarial sample when it can recognize every adversarial word embedding correctly. 
We summarize our contributions in the following:

\begin{itemize}
\setlength{\itemsep}{0pt}
    \item We design a contrastive adversarial representation learning approach to learn adversarial examples in the representation space, which can directly improve the encoder's robustness. 
    \item We propose a novel adversarial training task, RAR (Reconstruction from Adversarial Representations), to help the model learn a more robust representation at the word level.
    \item We conduct experiments on four text classification datasets and results show that our proposed approaches outperform strong baseline on accuracy and robustness. We release the source code at a GitHub repo.\footnote{https://github.com/FFYYang/CARL\_RAR}
\end{itemize}

\section{Related Work}

\paragraph{Gradient-based Adversarial Training.}
Adversarial examples were explored primarily in the computer vision area and received more attention in natural language processing recently. Different from the CV domain, we can improve NLP models' robustness and performance at the same time \cite{miyato2016adversarial}. 
\newcite{miyato2016adversarial} proposed to add perturbations calculated from gradient on word embeddings to obtain adversarial samples in embedding space.
\newcite{madry2017towards} proposed the k-PGD method and calculated adversarial perturbations through multiple forward-backward iterations to avoid the obfuscated gradient problem. It is widely accepted as the most effective approach, but multiple iterations leads to high computation cost. To mitigate the cost, \newcite{zhang2019you} restricted most perturbation updates in the first layer. \newcite{shafahi2019adversarial} designed a "free" algorithm that simultaneously updates both model parameters and adversarial perturbations in a single backward pass. \newcite{zhu2019freelb} proposed FreeLB which simultaneously accumulates the “free” parameter gradients in each iteration and updates the model parameters all at once after all iterations.

\paragraph{Contrastive Learning.}
Contrastive learning has recently become a dominant component in self-supervised learning methods for computer vision, natural language processing (NLP). The goal of contrastive learning is to learn a representation that is close in a certain metric space for pairs with the same label, while push apart the representation between pairs with different labels  \cite{tian2019contrastive}. This method has been successfully used in recent years for representation learning and knowledge distillation. In this work, we apply it into the adversarial training by narrowing the representations of the adversarial sample and its corresponding original sample, while enlarging their distance from samples that belong to different classes.

\paragraph{Auto-Encoder.}
The auto-encoder \cite{rumellhart1986learning} consists of two modules: the encoder and the decoder. The encoder is used to map the input sample $x$ to the feature space $z$, i.e. the encoding process. Then the abstract feature $z$ is mapped back to the original token space through a decoder to obtain the reconstructed sample $x'$, i.e. the decoding process. The optimization goal is to optimize both encoder and decoder by minimizing the reconstruction error, to learn the abstract feature representation $z$ for the input $x$. 

\begin{figure*}[htb]
\centering  
\subfigure[CARL architecture]{
\label{Fig.sub.1}
\includegraphics[width=0.55\textwidth]{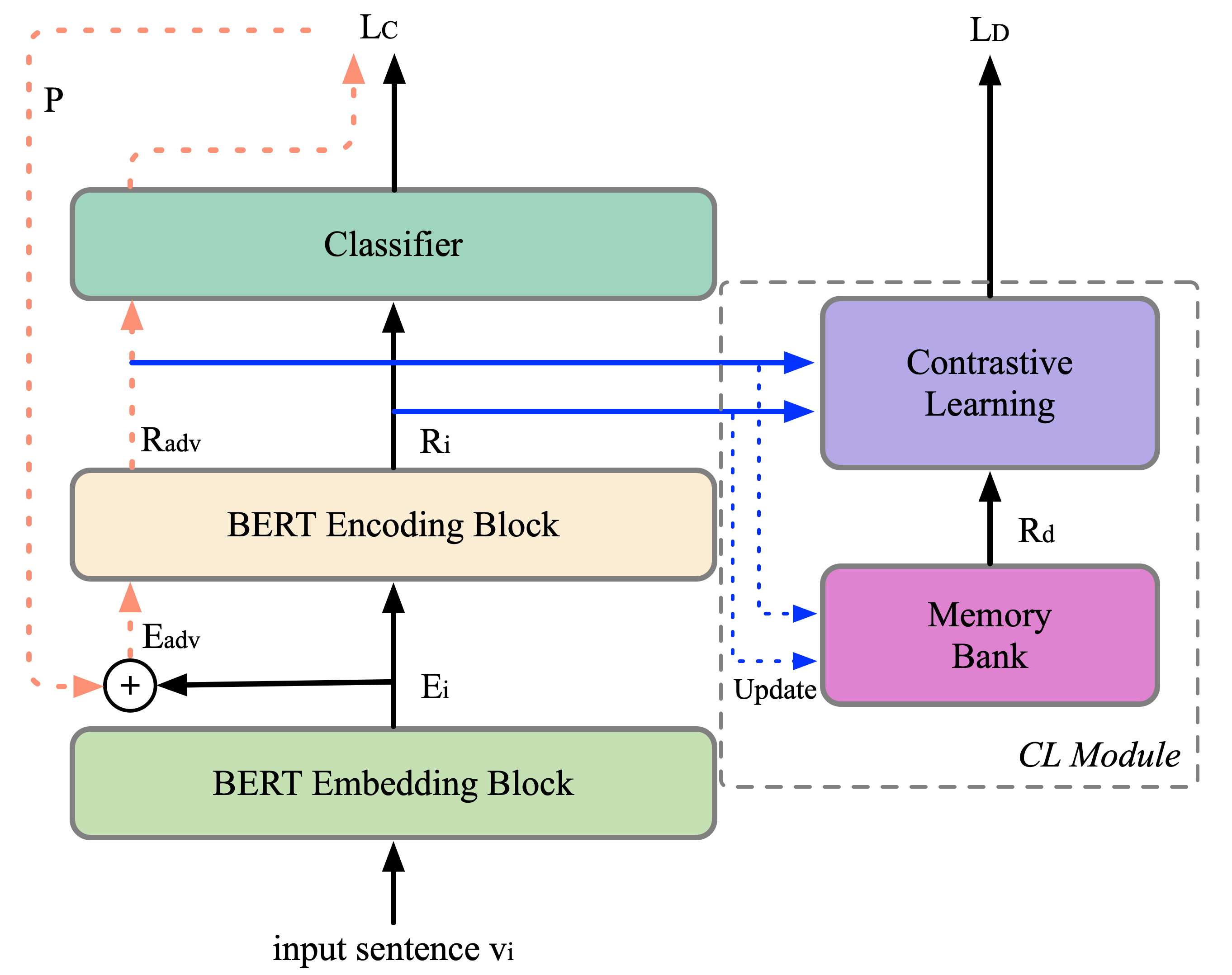}}
\subfigure[RAR architecture]{
\label{Fig.sub.2}
\includegraphics[width=0.35\textwidth]{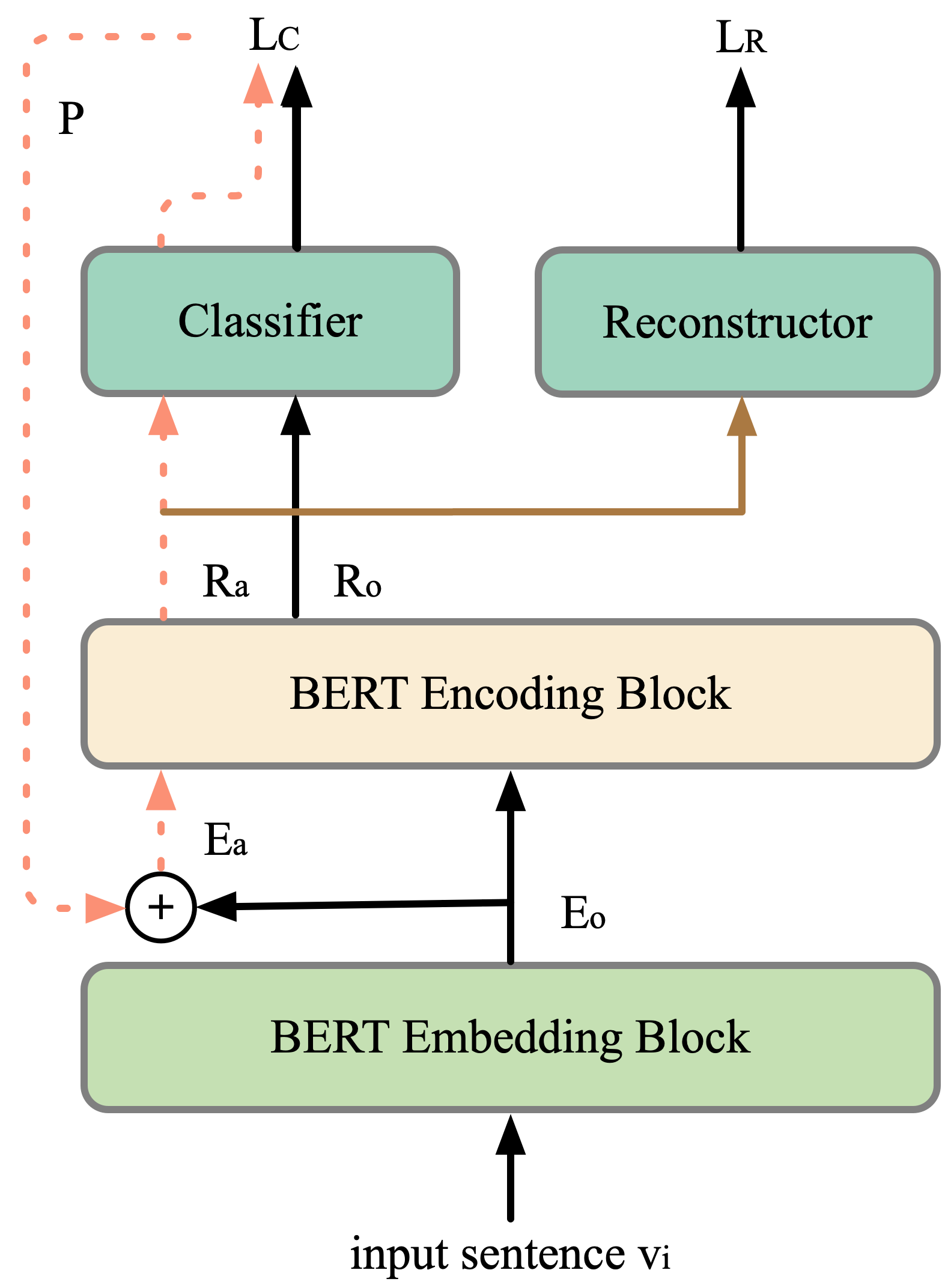}}
\caption{The overview of CARL and RAR. For each input training sentence, FreeLB is first used to generate its adversarial examples $E_a$, shown as the yellow dotted line. CL Module(\textbf{Left}) is used to calculate the contrastive loss which aims to narrow $R_a$ and $R_o$, while push them apart from $R_d$. Reconstructor(\textbf{Right}) is used to reconstruct the original sentence from $R_a$.}
\label{Fig.1}
\end{figure*}

In our work, we focus on the gradient-based adversarial training on the text classification where the model receives a sentence and outputs a single label. Though some neural networks have achieved promising results, they are vulnerable to the simple adversarial perturbations \cite{huang2017adversarial, yuan2019adversarial}. Some gradient-based adversarial training approaches were proposed to solve this problem \cite{zhu2019freelb, shafahi2019adversarial,madry2017towards, miyato2016adversarial}. Most of them focus on the generation of adversarial examples, but we focus on how to use these examples to train the model more efficiently by combining the idea of contrastive learning and auto-encoder.

\section{Approach}
We aim to learn a robust text classification model by helping the model to learn adversarial samples more efficiently in the training process.

\subsection{Overview}
The overview of our approaches is depicted in Figure \ref{Fig.1}.
Given an input training sentence, we first use FreeLB \cite{zhu2019freelb} to get its adversarial embeddings $E_a$ which are likely to fool the current model.
In addition to minimizing these adversarial examples' classification errors, we propose two novel approaches to train them: 
\textbf{1) CARL (Contrastive Adversarial Representation Learning).}
Its goal is to narrow the distance of sentence-level semantic representation between the original sample and its adversarial sample while pushing them away from samples that belong to different classes. We achieve this by using the CL (Contrastive Learning) module shown in Figure \ref{Fig.sub.1}.
\textbf{2) RAR (Reconstruction from Adversarial Representations).}
It is designed to reconstruct every word in the original input sentence from their adversarial representations by the reconstructor shown in Figure \ref{Fig.sub.2}.

In subsequent sections, we describe how to use CARL and RAR to train adversarial samples more effectively. In section 3.2, we describe how to use contrastive learning approach to learn a robust semantic representation for the input sentence. In section 3.3, a reconstruction module is designed to prompt the model to learn more robust lexical knowledge from input sentence's adversarial embeddings.

\subsection{Contrastive Adversarial Representation Learning}

\paragraph{Intuition.}
Recent gradient-based adversarial training approaches only use the classification loss to optimize the model on adversarial examples. Although they get promising results, the potential value of adversarial examples is not fully exploited. When only the classification loss is used, the model tends to learn a robust classifier, the robustness of the feature encoder is not greatly improved. 
After all, the classification loss function does not explicitly force the model try to learn a representation which is robust to adversarial perturbations.

Representation knowledge is highly structured, because dimensions contain complex interdependencies \cite{tian2019contrastive}. If the model learns the adversarial samples in this perspective, there will be a huge learning space. In addition, it is suitable for adversarial training, for the representation reflects the model's understanding and the extracted knowledge of the input sentence, which should be consistent, no matter the input is the original sentence or the adversarial sentence. 
We expect the model to directly learn an encoder which can output a robust semantic representation for the input sentence, and even if the input sentence is contaminated by adversarial perturbations, the representation will not be significantly affected. 

\begin{figure}[]
\centering  
\includegraphics[width=0.4\textwidth]{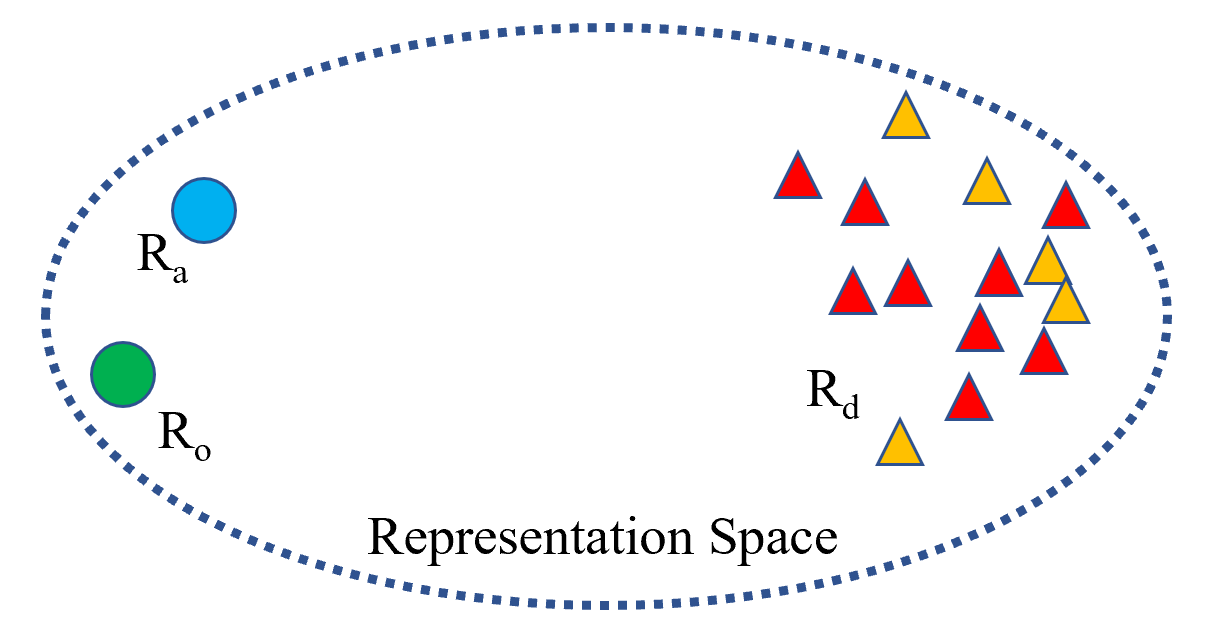}
\caption{The intuition of CARL. The blue and green circle are adversarial and original representation of the input example, the triangles are representations of examples which belong to different classes. CARL aims to get two circles close and keep circles away from triangles}
\label{Fig.2}
\end{figure}

The intuition of CARL is shown in Figure 2. The big ellipse refers to the representation space which is corresponding to the output of \textit{ALBERT Encoding Block}. 
$R_{o}$, shown as the small green circle, is the representation of the original training example. $R_{a}$, shown as the small blue circle, is the representation of adversarial example.
$R_{d}$, shown as small triangles, is a group of the representations of examples whose golden labels are different from the input sentence.
CARL's goal is to make two small circles closer and make circles far away from triangles, so as to prevent adversarial attacks from leading the model to incorrectly understand the input sentence. 

\paragraph{Implementation.}
We are inspired by the contrastive representation distillation approach proposed by \newcite{tian2019contrastive} and we adapt it to the text domain's adversarial training. Concretely, we design CARL's objective to maximize the lower bound to the mutual information between the adversarial and original representation of the input sentence. 

Specifically, given a dataset $V$ that consists of a collection of samples $\{v_i \}_{i=1}^N$. For each sample $v_i$, there are many other samples that share the same label with it, we call these samples \textit{positives}, accordingly, we call samples whose labels are different from $v_i$ as \textit{negatives}. In addition, the adversarial sample of $v_i$ can also be called \textit{positive}.

During model's training process, for each input sample $v_i$ whose embedding is $E_i$, we sample $K$ \textit{negatives} $\{v_{i, j}^n \}_{j=1}^K$ for it. FreeLB algorithm is first used to obtain a perturbation $\delta$ which can approximately maximum classification loss inside the $\epsilon$-ball around $E_i$, as
\begin{equation}
\max_{||\delta||<\epsilon} L_C(f_\theta(E_i + \delta), y_i),
\end{equation}
where $y_i$ is the golden label of $v_i$, $L_C$ is the classification loss function, $\theta$ is model's parameter, $f$ is the model's forward function. Adding $\delta$ to $E_i$ can obtain the adversarial embedding $E_i^{adv}$. Model's encoding block will then map $E_i$ and $E_i^{adv}$ to the representation space to get $R_i$ and $R_i^{adv}$. Similarly, we can also get the original and adversarial representations for the negatives $\{v_{i, j}^n \}_{j=1}^K$, we mark them as $\{R^n_{i,j}\}_{j=1}^K$ and $\{R_{i,j}^{n,adv}\}_{j=1}^K$.
We expect the distance between $R_i$ and $R_i^{adv}$ to be as close as possible while pushing the representations of \textit{negatives} away from them.
To achieve this, we adapt the contrastive objective proposed by \newcite{tian2019contrastive} into our optimization problem, as
\begin{equation}
\begin{split}
&L_{D}^{a} = - \mathop{E}\limits_{S_{adv}} \left[ \log{ \frac{h_\theta(\{R_i^{adv}, R_i \})} { \sum_{j=1}^K h_\theta(\{R_i^{adv}, R^n_{i,j}\}) }  } \right], \\
&L_{D}^{o} = - \mathop{E}\limits_{S_{orig}} \left[ \log{ \frac{h_\theta(\{R_i, R_i^{adv} \})} { \sum_{j=1}^K h_\theta(\{R_i, R_{i,j}^{n,adv}\})} } \right], \\
\end{split}
\end{equation}
\begin{equation}
L_{D} = L_{D}^{o} + L_{D}^{a},
\end{equation}
where $L_D^{a}$ is the contrastive loss function anchored on the adversarial representation $R_i^{adv}$ of $v_i$, it aims to force the input sentence's original representation and adversarial representation close, and push the adversarial representation of the input sentence apart from its \textit{negatives}' original representations, and it is optimized on set $S_{adv} = \{ R_i^{adv}, R_i, R^n_{i,1},...,R^n_{i,K}\}$. Similarly, $L_D^{o}$ is anchored on the original representation $R_i$ of $v_i$ and $s_{orig} = \{ R_i, R_i^{nadv}, R_{i,1}^{n,adv},...,R_{i,K}^{n,adv}\}$.
$h_\theta$ is a discriminating function which outputs a big value for \textit{positive} pairs and small for \textit{negative} pairs, we use vector dot product's result as the score and adjust its dynamic range by a hyperparameter $\tau$, as
\begin{equation}
h_\theta({x_1, x_2}) = \exp(x_1 \cdot x_2 \cdot \frac{1}{\tau}).
\end{equation}

In practice, $K$ can be extremely large. To make the computation of Eq.2 tractable, we randomly select $m(m<K)$ negatives from the dataset. Besides, Noise-Contrastive Estimation \cite{gutmann2010noise, wu2018unsupervised} is used to approximate the softmax distribution as well as reduce the computational cost. 

During the model's training process, for every training sample, we need $m$ negatives' original and adversarial representations. For $m$ is usually large in practice, so it is impossible to calculate all of these representations at the same time during each mini-batch's iteration. Following \newcite{wu2018unsupervised}, we maintain two memory banks, $B_{orig}, B_{adv}$, to store the original and adversarial representations for every training sample. Therefore, when we calculate the contrastive loss, we don't have to recompute \textit{negatives}' representations and we can just retrieve them from the memory bank. Besides, the memory bank should be dynamically updated with newly computed representations at each mini-batch iteration, as 
\begin{equation}
\begin{split}
&B_{orig}[i] = M \cdot B_{orig}[i] + (1 - M) \cdot R_i, \\
&B_{adv}[i] = M \cdot B_{adv}[i] + (1 - M) \cdot R_i^{adv},
\end{split}
\end{equation}
where $M$ is a hyperparameter, $i$ is the index of a training sample. To be noticed, CARL cannot be used at the beginning of training, because the model is unstable and both original and adversarial representations are noisy. Optimizing contrastive loss at this time can cause the model difficult to converge. The proper way is to wait until the model is going to be stable, and use an entire epoch to forward every training sample through the model to initialize the whole memory bank, after which the contrastive loss can be used to optimize the model. In conclusion, we will optimize the following problem, as 
\begin{equation}
\min\limits_{\theta} (L_C + L_D) _{(v,y)\sim D} \left[ \max\limits_{||\delta||<\epsilon} L_C(f_\theta(E + \delta), y) \right],
\end{equation}
where $v$ is one training sample, $y$ is its golden label, $D$ is the data distribution, $L_C$ is the classification loss.

\subsection{Reconstruction from Adversarial Representations}

\paragraph{Intuition.}
The gradient-based adversarial attacking approach adds perturbations on every word's embedding, we have no idea the contaminated embedding indicates which word in the real world. If the model cannot recognize the contaminated word embedding or identify it to a wrong word, its understanding of the whole sentence's semantics could be wrong, especially when the keyword of the sentence is misunderstood by the model. The special cases are easy to occur because we find that the norm of adversarial perturbation added to the keyword of a sentence is usually larger than that of others words, and it makes the keyword harder to be recognized.

To solve the problem, inspire by the Masked Language Model proposed in BERT \cite{devlin2018bert}. we design RAR to reconstruct every token from its adversarial representation. To reconstruct tokens correctly, the model should not only learn more robust lexical knowledge for every word but also accurately understand the semantics of the whole sentence.

\paragraph{Implementation.}
Inspired by the pre-training task used in BERT\cite{devlin2018bert}, we map the adversarial representation of each word to a vector which length is the vocabulary size. 

Specifically, the reconstructor receives input sentence's token-level adversarial representation $R_i^{adv, tok} \in [sequence\_length, hidden\_size]$ from the \textit{ALBERT Encoding block} as input, then $R_i^{adv, tok}$ will be forwarded through a \textit{Layer Normalization}, \textit{GeLU Activation Function} and two \textit{Feed Forward Layers}. 
The first feed-forward layer maps the $hidden\_size$ to $embedding\_size$ and the second feed-forward layer's parameters are shared with \textit{ALBERT Embedding Layer} to project the $embedding\_size$ into $vocabulary\_size$. Then, we can get the predicted probability distribution over the vocabulary for every token's position in the sentence. Finally, we use the cross-entropy function to calculate the reconstruction loss $L_{R}$.

In the training process, FreeLB and RAR are combined to optimize the model. After we use FreeLB to get the adversarial representations of every word and the whole sentence, we simultaneously feed them to the reconstructor and the classifier accordingly. That is, the model is asked not only to predict the correct class of the adversarial sample but also to reconstruct the sample's original words from their adversarial representations. 
In conclusion, we will optimize the following problem, as 
\begin{equation}
\small
\min\limits_{\theta} (L_C + w_r \cdot L_R) _{(v,y)\sim D} \left[ \max\limits_{||\delta||<\epsilon} L_C(f_\theta(E + \delta), y) \right],
\end{equation}
where $w_r=0.1$ is the weight for the reconstruction loss.

\section{Experiment}
We evaluate our approaches on four datasets. We first introduce the datasets, the baselines, and the experiment settings. Then, we show experiment results and provide further analysis.

\subsection{Datasets}

We use four text classification datasets: SST-2, Yelp-P, AG's News, and Yahoo! Answers.

\paragraph{SST-2.}
The Stanford Sentiment Treebank \cite{socher2013recursive} consists of sentences from movie reviews and human annotations of their sentiment. The task is to predict the sentence-level sentiment (positive/negative) of a given input text.

\paragraph{Yahoo! Answers.} 
This dataset is composed of ten topic categories: Society \& Culture, Science \& Mathematics, Health, Education \& Reference, etc. In this work, we use five categories. For every category, we use 12,000 training samples, 400 validation, and 400 test samples.

\paragraph{Yelp-P.}
The original Yelp dataset is built using reviews from the website Yelp\footnote{https://www.yelp.com/dataset/challenge}. Each review has a rating label varying from 1 to 5. We use it as the binary classification, and randomly choose 30,000 training samples, 1000 validation, and 1000 testing samples for every class.

\paragraph{AG's News.}
This is a dataset of more than one million news articles and they are categorized into four classes: World, Sports, Business, and Sci/Tech. Each class contains 30,000 training samples and 1,900 testing samples. In our work, for each class, we use 15,000 training samples, 500 validation and testing samples.

\subsection{Baselines}

We compare our proposed approach with the following approaches.

\paragraph{ALBERT for Text Classification.}
For ALBERT, the first token of the sequence is $[CLS]$, when doing the text classification task, ALBERT takes the final hidden state $h$ of the $[CLS]$ token as the representation of the whole sentence. The classifier consists of a feed-forward layer and a softmax function.
\begin{equation}
p(c|h) = softmax(Wh), 
\end{equation}
where $W$ is a learnable parameter matrix, $c$ is the class. ALBERT is fine-tuned with all parameters as well as $W$ jointly by maximizing the log-probability of the golden label.

\paragraph{FreeLB.}
FreeLB, proposed by \newcite{zhu2019freelb}, adds adversarial perturbations to ALBERT embedding layer's output, and minimizes the resultant adversarial loss around input samples, it leverages the "free" training strategy \cite{shafahi2019adversarial} to improve the efficiency of adversarial training, which made it possible to apply PGD-based adversarial training\cite{madry2017towards} into large-scale pre-trained language model.
In this work, we apply FreeLB to ALBERT model.

\begin{table}
\centering
\begin{tabular}{lllll}
\hline
\textbf{   } & \textbf{SST-2} & \textbf{Yahoo!} & \textbf{Yelp-P} & \textbf{AG's News} \\
\hline
$\gamma$ & 0.6 & 0 & 0.5 & 0  \\
$\alpha$ & 0.1 & 0.01 & 0.05 & 0.01  \\
$\epsilon$ & 0 & 0 & 0 & 0 \\
$n$ & 2 & 3 & 3 & 3  \\
\hline
\end{tabular}
\caption{\label{FreeLB_Hyper}
Hyperparameters for FreeLB on 4 datasets: step size $\alpha$, maximum perturbation norm $\epsilon$ (if it is set to zero, the perturbation's norm is not limited), number of iteration steps $n$, magnitude of initial random perturbation $\gamma$.
}
\end{table}

\begin{table*}
\centering
\begin{tabular}{lllll}
\hline
\textbf{    } & \textbf{SST-2} & \textbf{Yahoo! Answers} & \textbf{Yelp-P} & \textbf{AG's News}\\
\hline
ALBERT          & 92.16 & 73.93 & 93.55 & 89.90 \\
FreeLB          & 93.23 & 74.28 & 93.93 & 90.85 \\
CARL(ours) & \textbf{93.77(+0.54)} & 74.65(+0.37) & \textbf{94.55(+0.62)} & \textbf{92.05(+1.20)}\\
RAR(ours) & 93.73(+0.50) & \textbf{74.88(+0.60)} & 94.4(+0.47) & 91.75(+0.90) \\
\hline
\end{tabular}
\caption{\label{Main_results}
Comparisons between CARL, RAR, and baselines on four datasets. ALBERT is the model trained without any adversarial training approach. FreeLB uses classification loss to learn adversarial examples. CARL and RAR are implemented based on FreeLB, they use additional optimization objectives for adversarial examples. We compare them with FreeLB and find CARL performs best in most cases.
}
\end{table*}

\begin{table}
\centering
\begin{tabular}{lllll}
\hline
\textbf{   } & \textbf{SST-2} & \textbf{Yahoo!} & \textbf{Yelp-P} & \textbf{AG's News} \\
\hline
$\tau$ & 6315 & 7200 & 5625 & 7750  \\
\hline
\end{tabular}
\caption{\label{CARL_start_steps}
Steps after which $L_D$ will start to be used in CARL before which only $L_C$ is used to optimize the model's parameters.
}
\end{table}

\subsection{Experiment settings}

We implement our two approaches on albert-base-v2 (from huggingface's pytorch implementation\footnote{https://github.com/huggingface/transformers}), the parameters of \textit{ALBERT Embedding Block} and \textit{ALBERT Encoding Block} are loaded from the pre-trained model, we do experiments on the fine-tuning stage.
We use the Adam optimizer to train the modules and the learning rate is set to 1e-5, and batch size is 16 for AG's News and 32 for the other three datasets. Since FreeLB's hyperparameters highly depend on the characteristic of the dataset, 
we apply hyperparameter search to every dataset and the searching results are shown in Table~\ref{FreeLB_Hyper}.
These hyperparameters stay unchanged in CARL and RAR. We train our models on two Tesla P40s.

CARL and RAR are both implemented based on the FreeLB. In RAR, $L_R$ is used to update the model's parameters from the beginning of the training. While in CARL, $L_D$ is used after the model is about to be stable (specific settings can be found in Table~\ref{CARL_start_steps}). Besides, $m$ is set to 20000 for YelpP and 16000 for the other three datasets. $\tau$ and $M$ is set to 0.07 and 0.5 
respectively. 
For SST-2, we use a development set to do the evaluation. To make the results reliable, we run each experiment three times with the same hyperparameters but different random seeds and report their average scores. For the other three datasets, we use a development set to choose the best training checkpoint and evaluate it on the test set.

\subsection{Results and Discussion}

The results of the proposed approach and baselines are shown in Table~\ref{Main_results}.
FreeLB, CARL, and RAR let the adversarial samples participate in the model's training process, so it's not surprising that all of them perform better than ALBERT. These improvements can be mainly attributed to the effect of data augmentation. 

The experiment results also show that the performance of CARL and RAR on four data sets is higher than FreeLB.
These results demonstrate that the approaches we proposed to defend against gradient-based adversarial attacks during the training process are effective and well applied to various text classification datasets.
We conjecture that this is because the contrastive objective can encourage the model to discover the true underlining knowledge which can determine the classification label from adversarial and original representation. This underline knowledge is robust against adversarial perturbation added on the original sample and won't be changed by modifying the statement of the sentence. 
When the model can learn this knowledge, its generalization and robustness will be improved.

When comparing CARL and RAR, CARL performs better than RAR in most cases. It is because CARL's training objective is to narrow the distance between the adversarial sample and the original sample in the representation space, while the classifier of the model is also based on the representation of the sentence, so the objective of CARL has a more straight forward contribution to the classification task than that of the RAR.

\begin{table}
\centering
\begin{tabular}{lllll}
\hline
\textbf{  Cosine } & $\alpha$=0.1 & $\alpha$=0.075 \\
\hline
ALBERT      & 0.851 & 0.871 \\
FreeLB      & 0.899 & 0.918 \\
CARL   & 0.917(+0.018) & 0.934(+0.016)\\
RAR   & \textbf{0.926(+0.027)} & \textbf{0.941(+0.023)}\\
\hline
\end{tabular}

\begin{tabular}{lllll}
\hline
\textbf{ Euclidean } & $\alpha$=0.1 & $\alpha$=0.075 \\
\hline
ALBERT      & 8.409 & 7.746 \\
FreeLB      & 6.477 & 5.776 \\
CARL   & 5.340(-1.137) & 4.668(-1.108)\\
RAR   & \textbf{5.121(-1.356)} & \textbf{4.453(-1.323)} \\
\hline
\end{tabular}
\caption{\label{Rep_distance}
The difference between original and adversarial representations of samples in AG's News test set. FreeLB, RAR, and CARL perform much better than ALBERT. We compare CARL and RAR with FreeLB, and we find RAR is the best.
}
\end{table}

\subsection{Analysis}
\paragraph{The difference between adversarial and original sample's representations.}
Table~\ref{Rep_distance} compares the Euclidean distance and cosine similarity between adversarial and original samples' sentence-level representations in four approaches. 
We use AG's News test set 
to do this experiment. We use the models trained by the above four approaches, and for every sample $v_i$, we first calculate its original representation $R_i$, and obtain their adversarial representation $R_i^{adv}$ using the k-PGD approach with the same hyperparameters setting, 
then measure their distance by the cosine similarity and the Euclidean distance. 
We also compare results when using different max perturbation norms $\alpha$ in k-PGD. The final result is the average of all samples.

Experiment results show that FreeLB, CARL, and RAR perform much better than ALBERT either on the cosine similarity or Euclidean distance, this indicates that the robustness of the model in the representation space can be effectively improved by optimizing the classification error of adversarial samples.
In addition, when compared with FreeLB, CARL, and RAR, the performance of RAR is the best, followed by CARL. This shows that our approaches are effective to further improve model representation space's robustness and RAR is more effective. 
The reason why RAR is better than CARL can be explained that the objective of RAR is more difficult than that of CARL. The optimization objective of RAR is at the token level, while CARL is at the sentence level, so RAR can encourage the model to learn additional lexical knowledge which is also beneficial for improving the semantic representation of the whole sentence.

\begin{table}
\centering
\begin{tabular}{lllll}
\hline
\textbf{ } & $\epsilon$=0.02 & $\epsilon$=0.075 & $\epsilon$=0.1 \\ 
\hline
ALBERT      & 86.6 & 72.60 & 70.2  \\
FreeLB      & 88.8 &   80.25 & 78.0 \\
CARL   & \textbf{90.0} & \textbf{81.4}   & \textbf{80.1}  \\
RAR   & 89.8 & 80.5 & 78.6 \\
\hline
\end{tabular}
\caption{\label{performance_robustness}
Performance robustness experiment results.
$\epsilon$ is the maximum perturbation norm.
CARL performs best under adversarial attacks of different strength.
}
\end{table}

\paragraph{The robustness of performance.}
We use the k-PGD method to attack models trained on AG's News by four approaches.
Experimental results showed that the performance of the FreeLB, CARL, and RAR is significantly better than ALBERT. That is because they allow the adversarial samples to participate in the model's training process. 
In the case of FreeLB, RAR, and CARL, CARL is the best, followed by RAR. The reason can be explained from the perspective of multi-task learning. If we regard CARL and RAR as two multi-task learning frameworks, it is obvious that compared to the reconstruction task used in RAR, the contrastive learning task used in CARL is more similar to the classification task, because both of these two tasks' objectives operate on sentence-level representations. 
In addition, RAR performs better on representation robustness while CARL performs better on performance robustness.
This indicates that although narrowing the representation distance between original and adversarial samples can improve the model's performance and robustness. It's not the case that the shorter distance, the more robust performance. 

\paragraph{Reconstructed adversarial samples.}
We let SST-2's dev set forward the trained RAR model and use the k-PGD method to attack it. Then we take the output logits of the RAR module to obtain the reconstructed sentence.
We find that we could get some text-form adversarial samples in this way.
The semantics of these reconstructed samples are almost identical with that of original samples, 
but they can fool the model trained by ALBERT successfully.
Table~\ref{Reconstruction_egs} shows some examples of the reconstructed sentences which can be used as text-form adversarial samples and can be further used as augmented data. 

\begin{table}
\centering
\begin{tabular}{ p{6.5cm} p{0.5cm} }
\hline
Outer-space buffs {\color{red}might} love this film, but others will find its pleasures {\color{red}intermittent}. & N\\
Outer-space buffs {\color{blue}would} love this film, but others will find its pleasures {\color{blue}occasional}. & P\\
\hline
The film {\color{red}will} play {\color{red}equally} well on {\color{red}both} the standard and giant screens. & P\\
The film {\color{blue}would} play {\color{blue}more} well on {\color{blue}all} the standard and giant screens. & N\\
\hline 
Why make a documentary about {\color{red}these} marginal historical figures & N\\
Why make a documentary about {\color{blue}the} marginal historical figures & P\\
\hline 
\end{tabular}
\caption{\label{Reconstruction_egs}
Reconstructed adversarial samples. The first line is the original sentence, the second line is the reconstructed sentence. N and P refers to negative and positive label the model predicted. The model can correctly classify the original sentences, but not these reconstructed sentences.
}
\end{table}

\section{Conclusion}
In this work, we propose two gradient-based adversarial training approaches, CARL and RAR, to improve the performance and robustness of text classification models. 
The key idea of CARL is narrowing the original sample and adversarial sample in the representation space.
While RAR forces the model to reconstruct the original tokens from their adversarial representations.
Experiments demonstrate our approaches outperform the baseline.  
The sentence representation and the model's performance are more robust, which proves the effectiveness of the proposed approaches.
Besides, RAR can be used to generate adversarial examples.

\section*{Acknowledgments}
We would like to thank all the reviewers for their insightful and valuable comments and suggestions.

\bibliographystyle{acl_natbib}
\bibliography{anthology,acl2021}

\begin{thebibliography}{23}
\expandafter\ifx\csname natexlab\endcsname\relax\def\natexlab#1{#1}\fi

\bibitem[{Devlin et~al.(2019)Devlin, Chang, Lee, and
  Toutanova}]{devlin2018bert}
Jacob Devlin, Ming-Wei Chang, Kenton Lee, and Kristina Toutanova. 2019.
\newblock \href {https://doi.org/10.18653/v1/N19-1423} {{BERT}: Pre-training of
  deep bidirectional transformers for language understanding}.
\newblock In \emph{Proceedings of the 2019 Conference of the North {A}merican
  Chapter of the Association for Computational Linguistics: Human Language
  Technologies, Volume 1 (Long and Short Papers)}, pages 4171--4186,
  Minneapolis, Minnesota. Association for Computational Linguistics.

\bibitem[{Gutmann and Hyv{\"a}rinen(2010)}]{gutmann2010noise}
Michael Gutmann and Aapo Hyv{\"a}rinen. 2010.
\newblock Noise-contrastive estimation: A new estimation principle for
  unnormalized statistical models.
\newblock In \emph{Proceedings of the Thirteenth International Conference on
  Artificial Intelligence and Statistics}, pages 297--304.

\bibitem[{Hjelm et~al.(2019)Hjelm, Fedorov, Lavoie-Marchildon, Grewal, Bachman,
  Trischler, and Bengio}]{hjelm2018learning}
R~Devon Hjelm, Alex Fedorov, Samuel Lavoie-Marchildon, Karan Grewal, Phil
  Bachman, Adam Trischler, and Yoshua Bengio. 2019.
\newblock \href {https://openreview.net/forum?id=Bklr3j0cKX} {Learning deep
  representations by mutual information estimation and maximization}.
\newblock In \emph{International Conference on Learning Representations}.

\bibitem[{Huang et~al.(2017)Huang, Papernot, Goodfellow, Duan, and
  Abbeel}]{huang2017adversarial}
Sandy~H. Huang, Nicolas Papernot, Ian~J. Goodfellow, Yan Duan, and Pieter
  Abbeel. 2017.
\newblock \href {https://openreview.net/forum?id=ryvlRyBKl} {Adversarial
  attacks on neural network policies}.
\newblock In \emph{5th International Conference on Learning Representations,
  {ICLR} 2017, Toulon, France, April 24-26, 2017, Workshop Track Proceedings}.
  OpenReview.net.

\bibitem[{Kim(2014)}]{kim2014convolutional}
Yoon Kim. 2014.
\newblock \href {https://doi.org/10.3115/v1/D14-1181} {Convolutional neural
  networks for sentence classification}.
\newblock In \emph{Proceedings of the 2014 Conference on Empirical Methods in
  Natural Language Processing ({EMNLP})}, pages 1746--1751, Doha, Qatar.
  Association for Computational Linguistics.

\bibitem[{Lai et~al.(2015)Lai, Xu, Liu, and Zhao}]{lai2015recurrent}
Siwei Lai, Liheng Xu, Kang Liu, and Jun Zhao. 2015.
\newblock Recurrent convolutional neural networks for text classification.
\newblock In \emph{Twenty-ninth AAAI conference on artificial intelligence}.

\bibitem[{Madry et~al.(2018)Madry, Makelov, Schmidt, Tsipras, and
  Vladu}]{madry2017towards}
Aleksander Madry, Aleksandar Makelov, Ludwig Schmidt, Dimitris Tsipras, and
  Adrian Vladu. 2018.
\newblock \href {https://openreview.net/forum?id=rJzIBfZAb} {Towards deep
  learning models resistant to adversarial attacks}.
\newblock In \emph{International Conference on Learning Representations}.

\bibitem[{Miyato et~al.(2017)Miyato, Dai, and
  Goodfellow}]{miyato2016adversarial}
Takeru Miyato, Andrew~M. Dai, and Ian~J. Goodfellow. 2017.
\newblock \href {https://openreview.net/forum?id=r1X3g2\_xl} {Adversarial
  training methods for semi-supervised text classification}.
\newblock In \emph{5th International Conference on Learning Representations,
  {ICLR} 2017, Toulon, France, April 24-26, 2017, Conference Track
  Proceedings}. OpenReview.net.

\bibitem[{Neekhara et~al.(2019)Neekhara, Hussain, Dubnov, and
  Koushanfar}]{neekhara2018adversarial}
Paarth Neekhara, Shehzeen Hussain, Shlomo Dubnov, and Farinaz Koushanfar. 2019.
\newblock \href {https://doi.org/10.18653/v1/D19-1525} {Adversarial
  reprogramming of text classification neural networks}.
\newblock In \emph{Proceedings of the 2019 Conference on Empirical Methods in
  Natural Language Processing and the 9th International Joint Conference on
  Natural Language Processing, {EMNLP-IJCNLP} 2019, Hong Kong, China, November
  3-7, 2019}, pages 5215--5224. Association for Computational Linguistics.

\bibitem[{Pang et~al.(2002)Pang, Lee, and Vaithyanathan}]{pang2002thumbs}
Bo~Pang, Lillian Lee, and Shivakumar Vaithyanathan. 2002.
\newblock \href {https://doi.org/10.3115/1118693.1118704} {Thumbs up? sentiment
  classification using machine learning techniques}.
\newblock In \emph{Proceedings of the 2002 Conference on Empirical Methods in
  Natural Language Processing, {EMNLP} 2002, Philadelphia, PA, USA, July 6-7,
  2002}, pages 79--86.

\bibitem[{Ren et~al.(2019{\natexlab{a}})Ren, Deng, He, and
  Che}]{ren2019generating}
Shuhuai Ren, Yihe Deng, Kun He, and Wanxiang Che. 2019{\natexlab{a}}.
\newblock Generating natural language adversarial examples through probability
  weighted word saliency.
\newblock In \emph{Proceedings of the 57th annual meeting of the association
  for computational linguistics}, pages 1085--1097.

\bibitem[{Ren et~al.(2019{\natexlab{b}})Ren, Deng, He, and
  Che}]{ren-etal-2019-generating}
Shuhuai Ren, Yihe Deng, Kun He, and Wanxiang Che. 2019{\natexlab{b}}.
\newblock \href {https://doi.org/10.18653/v1/P19-1103} {Generating natural
  language adversarial examples through probability weighted word saliency}.
\newblock In \emph{Proceedings of the 57th Annual Meeting of the Association
  for Computational Linguistics}, pages 1085--1097, Florence, Italy.
  Association for Computational Linguistics.

\bibitem[{Rumellhart(1986)}]{rumellhart1986learning}
DE~Rumellhart. 1986.
\newblock Learning internal representations by error propagation.
\newblock \emph{Parallel distributed processing}, 1:318--362.

\bibitem[{Shafahi et~al.(2019)Shafahi, Najibi, Ghiasi, Xu, Dickerson, Studer,
  Davis, Taylor, and Goldstein}]{shafahi2019adversarial}
Ali Shafahi, Mahyar Najibi, Mohammad~Amin Ghiasi, Zheng Xu, John Dickerson,
  Christoph Studer, Larry~S Davis, Gavin Taylor, and Tom Goldstein. 2019.
\newblock Adversarial training for free!
\newblock In \emph{Advances in Neural Information Processing Systems}, pages
  3353--3364.

\bibitem[{Socher et~al.(2013)Socher, Perelygin, Wu, Chuang, Manning, Ng, and
  Potts}]{socher2013recursive}
Richard Socher, Alex Perelygin, Jean Wu, Jason Chuang, Christopher~D Manning,
  Andrew~Y Ng, and Christopher Potts. 2013.
\newblock Recursive deep models for semantic compositionality over a sentiment
  treebank.
\newblock In \emph{Proceedings of the 2013 conference on empirical methods in
  natural language processing}, pages 1631--1642.

\bibitem[{Sun et~al.(2019)Sun, Qiu, Xu, and Huang}]{sun2019fine}
Chi Sun, Xipeng Qiu, Yige Xu, and Xuanjing Huang. 2019.
\newblock How to fine-tune bert for text classification?
\newblock In \emph{China National Conference on Chinese Computational
  Linguistics}, pages 194--206. Springer.

\bibitem[{Tian et~al.(2020)Tian, Krishnan, and Isola}]{tian2019contrastive}
Yonglong Tian, Dilip Krishnan, and Phillip Isola. 2020.
\newblock \href {https://doi.org/10.1007/978-3-030-58621-8\_45} {Contrastive
  multiview coding}.
\newblock In \emph{Computer Vision - {ECCV} 2020 - 16th European Conference,
  Glasgow, UK, August 23-28, 2020, Proceedings, Part {XI}}, volume 12356 of
  \emph{Lecture Notes in Computer Science}, pages 776--794. Springer.

\bibitem[{Wang et~al.(2018)Wang, Singh, Michael, Hill, Levy, and
  Bowman}]{wang2018glue}
Alex Wang, Amanpreet Singh, Julian Michael, Felix Hill, Omer Levy, and Samuel
  Bowman. 2018.
\newblock \href {https://doi.org/10.18653/v1/W18-5446} {{GLUE}: A multi-task
  benchmark and analysis platform for natural language understanding}.
\newblock In \emph{Proceedings of the 2018 {EMNLP} Workshop {B}lackbox{NLP}:
  Analyzing and Interpreting Neural Networks for {NLP}}, pages 353--355,
  Brussels, Belgium. Association for Computational Linguistics.

\bibitem[{Wu et~al.(2018)Wu, Xiong, Yu, and Lin}]{wu2018unsupervised}
Zhirong Wu, Yuanjun Xiong, Stella~X Yu, and Dahua Lin. 2018.
\newblock Unsupervised feature learning via non-parametric instance
  discrimination.
\newblock In \emph{Proceedings of the IEEE Conference on Computer Vision and
  Pattern Recognition}, pages 3733--3742.

\bibitem[{Xu et~al.(2019)Xu, Zhao, Yan, Zeng, Liang, and Xu}]{xu2019lexicalat}
Jingjing Xu, Liang Zhao, Hanqi Yan, Qi~Zeng, Yun Liang, and SUN Xu. 2019.
\newblock Lexicalat: Lexical-based adversarial reinforcement training for
  robust sentiment classification.
\newblock In \emph{Proceedings of the 2019 Conference on Empirical Methods in
  Natural Language Processing and the 9th International Joint Conference on
  Natural Language Processing (EMNLP-IJCNLP)}, pages 5521--5530.

\bibitem[{Yuan et~al.(2019)Yuan, He, Zhu, and Li}]{yuan2019adversarial}
Xiaoyong Yuan, Pan He, Qile Zhu, and Xiaolin Li. 2019.
\newblock Adversarial examples: Attacks and defenses for deep learning.
\newblock \emph{IEEE transactions on neural networks and learning systems},
  30(9):2805--2824.

\bibitem[{Zhang et~al.(2019)Zhang, Zhang, Lu, Zhu, and Dong}]{zhang2019you}
Dinghuai Zhang, Tianyuan Zhang, Yiping Lu, Zhanxing Zhu, and Bin Dong. 2019.
\newblock \href
  {http://papers.nips.cc/paper/8316-you-only-propagate-once-accelerating-adversarial-training-via-maximal-principle}
  {You only propagate once: Accelerating adversarial training via maximal
  principle}.
\newblock In \emph{Advances in Neural Information Processing Systems 32: Annual
  Conference on Neural Information Processing Systems 2019, NeurIPS 2019, 8-14
  December 2019, Vancouver, BC, Canada}, pages 227--238.

\bibitem[{Zhu et~al.(2020)Zhu, Cheng, Gan, Sun, Goldstein, and
  Liu}]{zhu2019freelb}
Chen Zhu, Yu~Cheng, Zhe Gan, Siqi Sun, Tom Goldstein, and Jingjing Liu. 2020.
\newblock \href {https://openreview.net/forum?id=BygzbyHFvB} {Freelb: Enhanced
  adversarial training for natural language understanding}.
\newblock In \emph{International Conference on Learning Representations}.

\end{thebibliography}

\appendix
\section{Appendices}
\label{sec:appendix}
We provide some details of experiment settings.

\subsection{Additional Experimental Details}

There is no significant difference in the training time between our proposed two approaches. For SST-2 and AG's News, it takes about two hours to train the model. For Yelp-P and Yahoo, it takes about ten hours.

The number of parameters in each model is shown in Table \ref{Parameters}. 
The number of parameters for ALBERT, FreeLB, and CARL is the same, while RAR has more parameters because there is an additional reconstructor module.

\begin{table}[h]
\centering
\begin{tabular}{ll}
\hline
\textbf{} & \textbf{\#Parameters} \\
\hline
\textbf{ALBERT}	& 11685122  \\
\textbf{FreeLB}	& 11685122 \\
\textbf{CARL}	& 11685122 \\
\textbf{RAR}	& 11813810 \\
\hline
\end{tabular}
\caption{\label{Parameters}
Number of parameters in each model.
}
\end{table}

\subsection{Hyperparameter Search Details}

Because the hyperparameters of FreeLB differ greatly in different datasets, we should search for the best hyperparameter configuration for each dataset. 
We first set the searching bounds of each hyperparameter as shown in Table \ref{Hyperparameter_Bounds}. Then we combine grid search and manual tuning approaches. Specifically, grid search is first used to search at a relatively large granularity, and then manual tuning is used to search at a small granularity. The criterion used for hyperparameter searching is the accuracy of the validation set. The searching result is also used in CARL and RAR.

\begin{table}[h]
\centering
\begin{tabular}{ll}
\hline
\textbf{\textbf{Hyperparameter}} & \textbf{Bounds} \\
\hline
\textbf{$\gamma$}	& [0, 0.8]  \\
\textbf{$\alpha$}	& [0.01, 0.2] \\
\textbf{$\epsilon$}	& [0, 0.5] \\
\textbf{$n$}	& [2, 4] \\
\hline
\end{tabular}
\caption{\label{Hyperparameter_Bounds}
Bounds for each hyperparameter: Step size $\alpha$, maximum perturbation norm $\epsilon$ (if it is set to zero, the perturbation's norm is not limited), number of iteration steps $n$, the magnitude of initial random perturbation $\gamma$.
}
\end{table}

\subsection{Datasets Details}

The statistics information of four datasets is shown in Table \ref{Dataset_Details}. Except SST-2, we only use a portion of data which is randomly selected from the original dataset because of the limitation of computing resource. Since our goal is not to reach the SOTA but to gain relative improvement of performance and robustness compared to FreeLB, dropping some training data won't affect it.

The data pre-processing approach is the same as huggingface's implementation\footnote{https://github.com/huggingface/transformers}. In addition, we randomly sample $m$ \textit{negatives} for each training example in CARL.

\begin{table}[h]
\centering
\begin{tabular}{llllll}
\hline
\textbf{Dataset}  & \textbf{\#Train} & \textbf{\#Dev} & \textbf{\#Test}\\
\hline
\textbf{SST-2}		& 67,349	& 872 & -  \\
\textbf{Yahoo! Answers}		& 60,000 & 60,000	& 2,000 \\
\textbf{Yelp-P}		& 60,000 & 60,000	& 2,000	\\
\textbf{AG's News}		& 60,000 & 60,000	& 2,000	\\
\hline
\end{tabular}
\caption{\label{Dataset_Details}
The statistics information of the four datasets we use.
}
\end{table}

\end{document}



\appendix
\section{Appendices}
\label{sec:appendix}
In this section, we provide and explain some details of experiment settings, hyperparameter searching and the datasets we use.

\subsection{Additional Experimental Details}

There is no significant difference in the training time between our proposed two approaches. For SST-2 and AG's News, it takes about two hours to train the model. For Yelp-P and Yahoo, it takes about ten hours.


The number of parameters in each model is shown in Table \ref{Parameters}. 
The number of parameters for ALBERT, FreeLB and CARL is the same, while RAR has more parameters because there is an additional reconstructor module.

\begin{table}[h]
\centering
\begin{tabular}{ll}
\hline
\textbf{} & \textbf{\#Parameters} \\
\hline
\textbf{ALBERT}	& 11685122  \\
\textbf{FreeLB}	& 11685122 \\
\textbf{CARL}	& 11685122 \\
\textbf{RAR}	& 11813810 \\
\hline
\end{tabular}
\caption{\label{Parameters}
Number of parameters in each model.
}
\end{table}



\subsection{Hyperparameter Search Details}


Because the hyperparameters of FreeLB differ greatly in different datasets, we should search the best hyperparameter configuration for each dataset. 
We first set the searching bounds of each hyperparameter as shown in Table \ref{Hyperparameter_Bounds}. Then we combine grid search and manual tuning approaches. Specifically, grid search is first used to search at a relatively large granularity and then manual tuning is used to search at a small granularity. The criterion used for hyperparameter searching is the accuracy on the validation set. The searching result are also used in CARL and RAR.

\begin{table}[h]
\centering
\begin{tabular}{ll}
\hline
\textbf{\textbf{Hyperparameter}} & \textbf{Bounds} \\
\hline
\textbf{$\gamma$}	& [0, 0.8]  \\
\textbf{$\alpha$}	& [0.01, 0.2] \\
\textbf{$\epsilon$}	& [0, 0.5] \\
\textbf{$n$}	& [2, 4] \\
\hline
\end{tabular}
\caption{\label{Hyperparameter_Bounds}
Bounds for each hyperparameter: Step size $\alpha$, maximum perturbation norm $\epsilon$ (if it is set to zero, the perturbation's norm is not limited), number of iteration steps $n$, magnitude of initial random perturbation $\gamma$.
}
\end{table}




\subsection{Datasets Details}

The statistics information of four datasets are shown in Table \ref{Dataset_Details}. Except SST-2, we only use a portion of data which is randomly selected from the original dataset because of the limitation of computing resource. Since our goal is not to reach the SOTA but to gain relative improvement of performance and robustness compared to FreeLB, dropping some training data won't affect it.

The data pre-processing approach is the same as huggingface's implementation\footnote{https://github.com/huggingface/transformers}. In addition, we randomly sample $m$ \textit{negatives} for each training example in CARL.

\begin{table}[h]
\centering
\begin{tabular}{llllll}
\hline
\textbf{Dataset}  & \textbf{\#Train} & \textbf{\#Dev} & \textbf{\#Test}\\
\hline
\textbf{SST-2}		& 67,349	& 872 & -  \\
\textbf{Yahoo! Answers}		& 60,000 & 60,000	& 2,000 \\
\textbf{Yelp-P}		& 60,000 & 60,000	& 2,000	\\
\textbf{AG's News}		& 60,000 & 60,000	& 2,000	\\
\hline
\end{tabular}
\caption{\label{Dataset_Details}
The statistics information of the four datasets we use.
}
\end{table}

